\pdfoutput=1

\documentclass[11pt]{article}

\usepackage[final]{acl}

\usepackage{times}
\usepackage{latexsym}

\usepackage[T1]{fontenc}

\usepackage[utf8]{inputenc}

\usepackage{microtype}

\usepackage{inconsolata}

\usepackage{graphicx}

\usepackage{booktabs}
\usepackage{pifont}
\usepackage{amsmath}
\usepackage{amssymb}
\usepackage{multirow}
\usepackage{xspace}
\usepackage{subcaption}

\newcommand{\cmark}{\ding{51}}%
\newcommand{\xmark}{\ding{55}}%
\newcommand{\task}[0]{Product Attribute Value Identification\xspace}
\newcommand{\taskave}[0]{Product Attribute Value Extraction\xspace}
\newcommand{\taskabbr}[0]{PAVI\xspace}
\newcommand{\taskabbrave}[0]{PAVE\xspace}
\newcommand{\platform}[0]{\textit{Xianyu}\xspace}
\newcommand{\method}[0]{TACLR\xspace}

\newcommand{\datasetxy}[0]{Xianyu-PAVI\xspace}
\newcommand{\datasetwdc}[0]{WDC-PAVE\xspace}

\usepackage{svg}

\title{TACLR: A Scalable and Efficient Retrieval-Based Method \\for Industrial Product Attribute Value Identification}

\author{
 \textbf{Yindu Su\textsuperscript{1,2}},
 \textbf{Huike Zou\textsuperscript{1}},
 \textbf{Lin Sun\textsuperscript{3}},
 \textbf{Ting Zhang\textsuperscript{2}},
 \textbf{Haiyang Yang\textsuperscript{1}},
 \textbf{Liyu Chen\textsuperscript{1}},
 \\
 \textbf{David Lo\textsuperscript{2}},
 \textbf{Qingheng Zhang\textsuperscript{1}},
 \textbf{Shuguang Han\textsuperscript{1}},
 \textbf{Jufeng Chen\textsuperscript{1}},
\\
 \textsuperscript{1}Xianyu of Alibaba \quad
 \textsuperscript{2}Singapore Management University \quad
 \textsuperscript{3}Hangzhou City University
\\
\texttt{\href{mailto:yindusu@foxmail.com}{yindusu@foxmail.com}},
\texttt{\href{mailto:tingzhang.2019@phdcs.smu.edu.sg}{tingzhang.2019@phdcs.smu.edu.sg}}
\\
\texttt{\href{mailto:qingheng.zqh@alibaba-inc.com}{qingheng.zqh@alibaba-inc.com}, \href{mailto:shuguang.sh@alibaba-inc.com}{shuguang.sh@alibaba-inc.com}}
}

\begin{document}
\maketitle

\begin{abstract}
\task (\taskabbr) involves identifying attribute values from product profiles, a key task for improving product search, recommendation, and business analytics on e-commerce platforms.
However, existing \taskabbr methods face critical challenges, such as inferring implicit values, handling out-of-distribution (OOD) values, and producing normalized outputs.
To address these limitations, we introduce Taxonomy-Aware Contrastive Learning Retrieval (\method), the first retrieval-based method for \taskabbr.
\method formulates \taskabbr as an information retrieval task by encoding product profiles and candidate values into embeddings and retrieving values based on their similarity. It leverages contrastive training with taxonomy-aware hard negative sampling and employs adaptive inference with dynamic thresholds.
\method offers three key advantages: (1) it effectively handles implicit and OOD values while producing normalized outputs; (2) it scales to thousands of categories, tens of thousands of attributes, and millions of values; and (3) it supports efficient inference for high-load industrial deployment.
Extensive experiments on proprietary and public datasets validate the effectiveness and efficiency of \method. Further, it has been successfully deployed on the real-world e-commerce platform \platform, processing millions of product listings daily with frequently updated, large-scale attribute taxonomies. 
We release the code to facilitate reproducibility and future research at \href{https://github.com/SuYindu/TACLR}{\texttt{https://github.com/SuYindu/TACLR}}.
\end{abstract}

\section{Introduction}

\begin{figure}[t]
    \centering
    \includegraphics[width=\columnwidth]{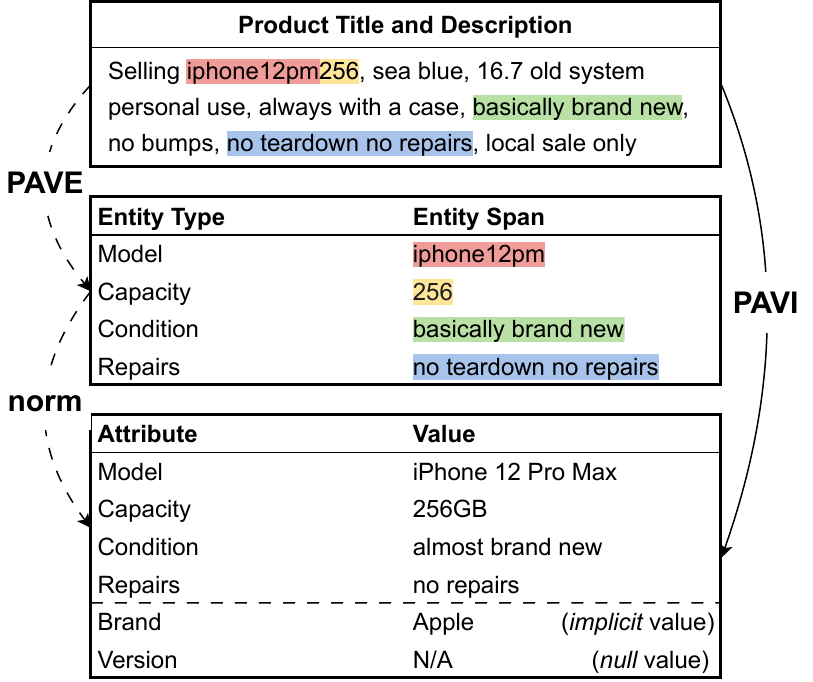}
    \caption{Illustration of the \taskabbrave\ and \taskabbr\ tasks. Unlike \taskabbrave, which extracts raw value spans from product profiles, \taskabbr\ requires outputs to be normalized and supports both the identification of \textit{implicit} values and the assignment of \textit{null} to unavailable attribute values.}
\label{fig:tasks}
\end{figure}

Product attribute values are key components that support the operation of e-commerce platforms. They provide essential structural information, aiding customers in making informed purchasing decisions and enabling product listing \cite{chen-etal-2024-ipl}, recommendation \cite{10.1145/3485447.3512018, 10.1145/3336191.3371805}, retrieval \cite{10.1145/3308560.3316603, 10.1145/2632165}, and question answering \cite{10.1145/3308560.3316597, 10.1145/3289600.3290992}.

However, seller-provided attribute values are often incomplete or even inaccurate—an issue that is particularly severe on second-hand e-commerce platforms such as \platform\footnote{\url{https://www.goofish.com}}.
This undermines the effectiveness of downstream applications, making the automatic identification of product attribute values a fundamental requirement.
Researchers have explored \taskave (\taskabbrave), which involves extracting spans from product profiles using Named Entity Recognition (NER) \cite{opentag} or Question Answering (QA) \cite{AVEQA} models. The upper part of Figure~\ref{fig:tasks} illustrates an example of NER-based \taskabbrave.

Although these approaches effectively extract value spans, the outputs remain raw text subsequences. Presenting attribute values in a standardized format is crucial for facilitating data aggregation in business analytics and enhancing the user experience by providing clear and consistent information.
To produce standardized values, a normalization step \cite{putthividhya-hu-2011-bootstrapped, 10.1145/3459637.3481946} is required to map these spans to predefined formats, as shown in the lower part of Figure~\ref{fig:tasks}. However, \textit{implicit} values, such as \texttt{Apple}, cannot be directly extracted and must instead be inferred from context or prior knowledge.

\begin{figure}[t]
    \centering
    \includegraphics[width=0.8\columnwidth]{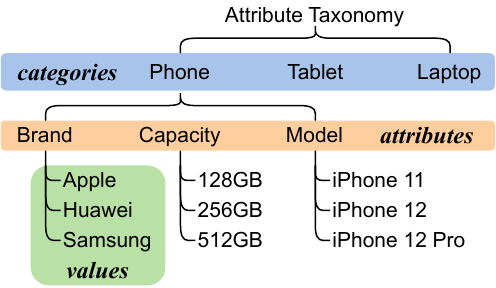}
    \caption{An illustration of a portion of the attribute taxonomy. Each category, such as \textit{Phone}, is linked to multiple attributes, including \textit{Brand}, \textit{Model}, and \textit{Capacity}, with standardized values enumerated for each attribute (e.g., \textit{Apple}, \textit{Huawei}, and \textit{Samsung} for \textit{Brand}).}
    \label{fig:taxonomy}
\end{figure}

Therefore, in this work, we focus on \task (\taskabbr) \cite{shinzato-etal-2023-unified}, which aims to associate predefined attribute values from attribute taxonomy (illustrated in Figure~\ref{fig:taxonomy}) with products. The input for \taskabbr includes the product category and profile, where the profile includes textual data, such as the title and description, and may optionally incorporate visual information, such as images or videos. The output is a dictionary with predefined attributes as keys and their inferred values as corresponding entries.
In addition, \taskabbr requires determining when attribute values are missing. For instance, as shown in Figure~\ref{fig:tasks}, the value for \texttt{Version} is unavailable and is therefore assigned an empty result or null value.

Beyond adapting \taskabbrave approaches, researchers have investigated classification-based~\cite{chen-etal-2022-extreme} and generation-based paradigms~\cite{sabeh2024empirical} for \taskabbr.
Classification-based methods treat each value as a class; while this approach is straightforward, it is fundamentally limited by the inability to identify out-of-distribution (OOD) values not present in the training data, making such methods impractical for the continuously evolving nature of e-commerce platforms.
In contrast, generation-based methods reformulate \taskabbr as a sequence-to-sequence task. Although these methods can handle implicit and OOD values, they face significant challenges, such as generating uncontrollable outputs and incurring substantial computational costs in high-load scenarios due to their reliance on Large Language Models (LLMs).
In summary, existing approaches face distinct challenges, including difficulties in identifying implicit values, generalizing to OOD values, producing normalized outputs, and ensuring scalability and efficiency for large-scale industrial applications.

To address these limitations, we propose a novel retrieval-based method, Taxonomy-Aware Contrastive Learning Retrieval (\method). Our approach formulates \taskabbr as an information retrieval task: the product item serves as the query, and the attribute taxonomy acts as the corpus, enabling efficient retrieval of relevant attribute values as matched documents.
We use a shared encoder to generate embeddings for both the input product and candidate values from the attribute taxonomy.
Our method adopts a contrastive learning framework inspired by CLIP \cite{pmlr-v139-radford21a}. Rather than relying on in-batch negatives, we implement taxonomy-aware negative sampling, which selects hard negative values from the same category and attribute to generate a more challenging and precise contrastive signal. Additionally, learnable null values are explicitly incorporated as the ground truth for product-attribute pairs without associated values.
During inference, we address the limitations of static thresholds by introducing dynamic thresholds derived from the relevance score of null values. This adaptive inference mechanism improves the model's ability to generalize across a large-scale attribute taxonomy.

Our contributions are threefold:
(1) We propose a novel retrieval-based paradigm for \taskabbr, introducing a scalable and efficient framework capable of handling implicit values, generalizing to OOD values, and producing normalized outputs.
(2) We incorporate contrastive training into \method{}, using a taxonomy-aware negative sampling strategy to improve representation discrimination, and introduce an adaptive inference mechanism that dynamically balances precision and recall in large-scale industrial applications.
(3) We validate the effectiveness of \method{} through extensive experiments on proprietary and public datasets, and demonstrate its successful deployment in a real-world industrial environment, processing millions of product listings across thousands of categories and millions of attribute values.

\section{Related Work}
\label{sec:related_work}


\subsection{Product Attribute Value Extraction}

\noindent \textbf{PAVE as Named Entity Recognition.} PAVE can be formulated as NER by identifying subsequences in product texts as entity spans and associating them with attributes as entity types. Early methods, such as OpenTag \cite{opentag}, trained individual models for each category-attribute pair. Subsequent efforts generalized this approach to support multiple attributes or categories. For instance, SUOpenTag \cite{xu-etal-2019-scaling} incorporated attribute embeddings into an attention layer to handle multiple attributes, while AdaTag \cite{yan-etal-2021-adatag} used attribute embeddings to parameterize the decoder. TXtract \cite{karamanolakis-etal-2020-txtract} introduced a category encoder and a category attention mechanism to tackle various categories effectively. Additionally, M-JAVE \cite{zhu-etal-2020-multimodal} jointly modeled attribute prediction and value extraction tasks while also incorporating visual information. More recently, \citet{chen-etal-2023-named} scaled BERT-NER by expanding the number of entity types to support a broader range of attributes.

\noindent \textbf{PAVE as Question Answering.} The QA framework can also be adapted for \taskabbrave by treating the product profile as context, attributes as questions, and value spans extracted from the context as answers. \citet{AVEQA} first introduced AVEQA for QA-based \taskabbrave. Subsequent work extended this framework by incorporating multi-source information \cite{MAVEQA}, multi-modal feature \cite{wang-etal-2022-smartave}, and trainable prompts \cite{yang-etal-2023-mixpave}. Moreover, the question can be extended by appending candidate values as demonstrated by \cite{shinzato-etal-2022-simple}. Combining NER and QA paradigms, \citet{ding-etal-2022-ask} proposed a two-stage framework, which first identifies candidate values and then filters them.

While NER- and QA-based paradigms have proven effective for PAVE, they struggle to identify implicit attribute values. Additionally, both paradigms rely on post-extraction normalization to standardize values, using either lexical \cite{putthividhya-hu-2011-bootstrapped} or semantic methods \cite{10.1145/3459637.3481946}.
Furthermore, extraction-based models require token-level annotations (e.g., BIO tags) for training and evaluation. Producing such annotations is significantly more resource-intensive than generating the value-level annotations used by \method, further limiting the scalability of these extraction-based methods.


\begin{table}[t]
\centering
\small
\caption{Comparison of different paradigms for identifying implicit, OOD, and normalized values.}
\begin{tabular}{lccc}
\toprule
Paradigm & Implicit & OOD & Normalized \\
\midrule
Extraction       & \xmark & \cmark & \xmark \\
Classification   & \cmark & \xmark & \cmark \\
Generation       & \cmark & \cmark & \xmark \\
\midrule
Retrieval        & \cmark & \cmark & \cmark \\
\bottomrule
\end{tabular}
\label{tab:paradigm_compare}
\end{table}

\subsection{Product Attribute Value Identification}

\noindent \textbf{Classification-Based \taskabbr.}
A straightforward approach is to frame \taskabbr as a multi-label classification problem over a finite set of values. \citet{chen-etal-2022-extreme} trained a unified classification model that masks invalid labels based on the product category.
However, a significant limitation of this classification-based paradigm is its inability to recognize OOD values not included in the training set. Consequently, its practicality is limited in dynamic e-commerce environments, where new categories and values frequently emerge.


\noindent \textbf{Generation-Based \taskabbr.}
Recent advancements in LLM have spurred the exploration of generation-based \taskabbr methods \cite{sabeh2024empirical}. Some methods \cite{roy-etal-2021-attribute, nikolakopoulos2023sage, blume-etal-2023-generative} construct attribute-aware prompts to generate values for each attribute individually. In contrast, others generate values for multiple attributes simultaneously, either in a linearized sequence format \cite{shinzato-etal-2023-unified} or as a hierarchical tree structure \cite{li-etal-2023-attgen}. Multimodal information has also been integrated into LLMs to identify implicit attribute values from product images \cite{PAM, khandelwal-etal-2023-large}. More recently, \citet{Baumann2024UsingLF} explored the use of LLMs for both the extraction and normalization of attribute values. Additionally, \citet{zou-etal-2024-eiven} introduced the learning-by-comparison technique to reduce model confusion, and \citet{sabeh2024exploringlargelanguagemodels} investigated Retrieval-Augmented Generation (RAG) technologies for \taskabbr.

Although generation-based methods can infer implicit and OOD attribute values from product profiles, they face several challenges in real-world scenarios. A key issue is the potential for the LLMs to produce uncontrollable or hallucinated outputs, a known limitation of LLMs \cite{10.1145/3703155}. Additionally, these methods often rely on large, computationally intensive models to achieve strong performance, making them inefficient and costly for large-scale industrial deployment.

\begin{figure*}[t]
  \centering
  \includegraphics[width=\linewidth]{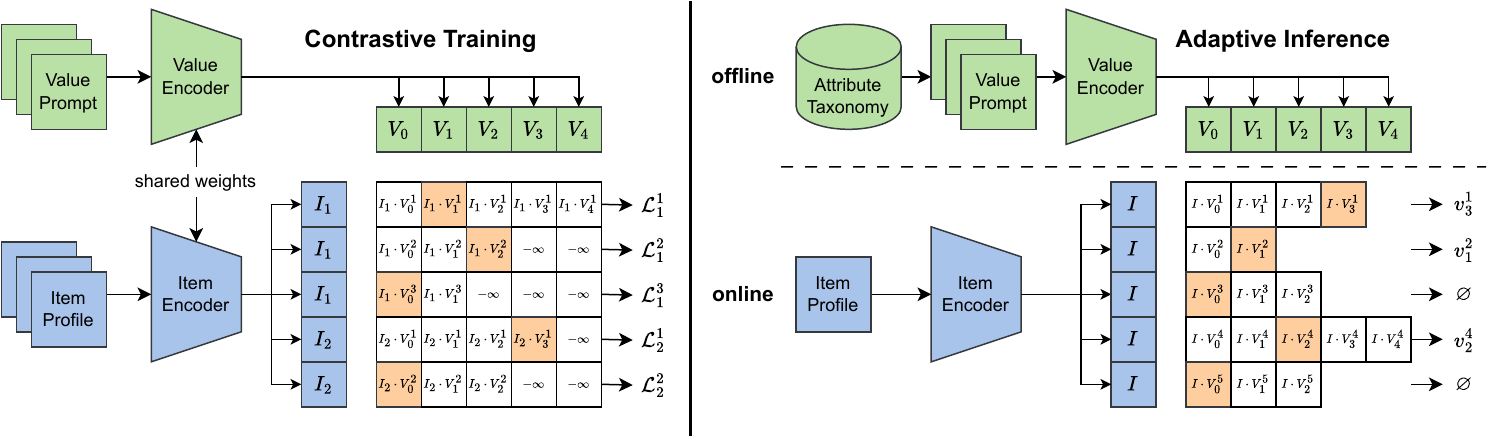}
    \caption{
    Overview of the training and inference process of TACLR, our retrieval-based method for the \taskabbr task.
    The left section illustrates contrastive training with taxonomy-aware negative sampling, while the right section demonstrates adaptive inference with pre-computed value embeddings.
    }
  \label{fig:method}
\end{figure*}

\section{Taxonomy-Aware Contrastive Learning Retrieval}

This section defines the \taskabbr task with an attribute taxonomy (\S\ref{sec:problem_def}) and presents our retrieval-based paradigm for \taskabbr (\S\ref{sec:retrieval-based}). We then detail the use of contrastive training with taxonomy-aware negative sampling (\S\ref{sec:contrastive_training}) and an adaptive inference mechanism with dynamic thresholds (\S\ref{sec:adaptive_inference}). Figure~\ref{fig:method} provides an overview of our approach \method.


\subsection{\taskabbr Task Definition}
\label{sec:problem_def}

\taskabbr is grounded in an attribute taxonomy that encompasses numerous product categories. For each category \(c\), the taxonomy specifies a set of attributes \(\mathcal{A}_c = \{a_1, a_2, \dots\}\) relevant to products in that category. For each attribute \(a \in \mathcal{A}_c\), it provides a predefined set of standard values \(\mathcal{V}_a = \{v_1, v_2, \dots\}\). Figure~\ref{fig:taxonomy} illustrates this structure\footnote{Our approach focuses on attribute value identification, leveraging an existing attribute taxonomy as input rather than constructing or updating the taxonomy itself. In the \platform platform, a dedicated team and supporting system are responsible for maintaining the taxonomy (i.e., ``attribute mining'').}.

For a given product item \(i\), with its title \(t\) and description \(d\), the item is assigned to a category \(c\) with associated attributes \(\mathcal{A}_c\). The objective of the \taskabbr task is to identify a relevant set of values \(\mathcal{V}_a^+ \subseteq \mathcal{V}_a\) for each attribute \(a \in \mathcal{A}_c\). The set \(\mathcal{V}_a^+\) can take one of three forms: a singleton (\(\{v\}\)), multiple values (\(\{v_1, v_2, \dots\}\)), or an empty set (\(\varnothing\)) if no information about \(a\) is available in the product profile. 
Notably, a standard value may not always appear verbatim as a text span in \(t\) or \(d\); it may instead be present in a paraphrased or synonymous form, which we refer to as an \textit{unnormalized} value (e.g., the standard value \texttt{iPhone 12 Pro Max} expressed as \texttt{iphone12pm} in Figure~\ref{fig:tasks}). In other cases, a value may not be explicitly mentioned in the product profile but can be inferred from the context; these are referred to as \textit{implicit} values (e.g., \texttt{Apple} in Figure~\ref{fig:tasks}).

\subsection{Retrieval-Based \taskabbr}
\label{sec:retrieval-based}

In a standard information retrieval setting, given a query, the objective is to retrieve a list of relevant documents from a corpus. Similarly, for \taskabbr, we treat the input item as the query and the attribute taxonomy as the corpus, aiming to retrieve relevant attribute values as the output documents.

\noindent\textbf{Encoding of items and values.}
We preprocess both item profiles and candidate values as textual inputs, utilizing a shared text encoder. 
For each item, we concatenate its title (\(t\)) and description (\(d\)) into a single input sentence formatted as: \texttt{title: \{title\} description: \{description\}}.
Each candidate value is represented as a context-rich prompt, structured as: \texttt{A \{category\} with \{attribute\} being \{value\}}, e.g., \texttt{A phone with brand being Apple}.
We explore the impact of various prompt templates in \S\ref{sec:analysis}
\footnote{This framework can be extended to multimodal scenarios by replacing the text encoder with a multimodal encoder to incorporate features such as images.}.

\noindent\textbf{Inference pipeline.}
During the deployment process, all value embeddings within the attribute taxonomy are pre-computed offline and indexed using the Faiss library\footnote{\url{https://github.com/facebookresearch/faiss}}.
During online inference, each item is encoded into an embedding, which is then compared against groups of indexed candidate value embeddings for various attributes. For each attribute \(a \in \mathcal{A}_c\), the top-$k$ most similar values are retrieved whose similarity scores exceed an adaptive threshold (\S\ref{sec:adaptive_inference}).

\subsection{Contrastive Training}
\label{sec:contrastive_training}

Inspired by CLIP \cite{pmlr-v139-radford21a}, we employ contrastive learning to train the shared encoder. Rather than relying on in-batch negatives, we compare each positive value with hard negative values from the same category and attribute in the taxonomy, providing a more challenging and precise training signal.

Formally, the subset of values matched with the item is referred to as the ground truth value set, \(\mathcal{V}_{a}^+ \subseteq \mathcal{V}_{a}\). If no matched values exist for a given attribute, i.e., \(\mathcal{V}_a^+ = \varnothing\), we assign a specific \textit{null value} \(v_0^a\) for this attribute as the positive value, i.e. \(v_a^+ = v_0^a\). Otherwise, a positive value is randomly drawn from the ground truth value set, i.e. \(v_a^+ \sim \mathcal{V}_a^+\).
For negative sampling, we select values as \( \mathcal{V}_a^- = \{v_1^{-}, v_2^{-}, \dots\} \subseteq \mathcal{V}_a - \mathcal{V}_a^+\), ensuring a maximum of \(K\) values. The contrastive loss is then computed as follows:
\[
\mathcal{L}_a = -\log\left(\frac{\exp(\frac{s(i, v_a^+)}{\tau})}{\exp(\frac{s(i, v_a^+)}{\tau}) + \sum\limits_{v \in \mathcal{V}_a^-} \exp(\frac{s(i, v)}{\tau})}\right)
\]
where \( s(i, v) = \frac{I \cdot V}{\|I\|\|V\|} \) denotes the cosine similarity between the item embedding \( I \) and the value embedding \( V \), and \( \tau \) is the temperature hyperparameter.
It is important to note that each item typically includes multiple attributes, all of which share the same item embedding \( I \) while being individually compared against corresponding values. Therefore, the loss for item \(i\) is the sum of losses over all attributes from \(\mathcal{A}_c\):
\[
\mathcal{L}_i = \sum_{a \in \mathcal{A}_c} \mathcal{L}_a
\]

An example logit matrix is depicted on the left side of Figure~\ref{fig:method}. Note that the item embedding \(I_1\) contributes to the loss computations of \(\mathcal{L}_1^1\), \(\mathcal{L}_1^2\), and \(\mathcal{L}_1^3\), which correspond to the attributes \(a_1\), \(a_2\), and \(a_3\) within the same product category.
We also pad the logit matrix with negative infinity for batched computation if fewer than \(K\) values are available.

\subsection{Adaptive Inference}
\label{sec:adaptive_inference}

During retrieval, relevance scores are assigned to every candidate values. To filter output values, a static threshold \( T \) can be applied to these scores. However, in real-world e-commerce platforms with a vast number of category-attribute pairs, using a single threshold across all pairs is often suboptimal. Moreover, defining a unique threshold for each pair is tedious or even impractical.

To address this, we introduce an adaptive inference method that uses dynamic thresholds to make cutoff decisions. As discussed in \S\ref{sec:contrastive_training}, we add an explicit null value \( v_0^a \) for each category-attribute pair, with its embedding learned during training. In the inference phase, we compute the similarity \( s(i, v_0^a) \) between the item and the null value, using it as a dynamic threshold \( T'_a \) to exclude candidate values for attribute \( a \) that have lower scores:
\[
\mathcal{V}_a^{\text{pred}} = \{v \mid s(i, v) > T'_a\}
\]
Since most category-attribute pairs have exclusive values, meaning that each product can have at most one value for a given attribute, we focus on the top-1 predicted value in this work. The output can be further simplified as follows:
\[
v_a^{\text{pred}} = 
\begin{cases} 
\arg\max\limits_{v \in \mathcal{V}_a} s(i, v) & \text{if } \max\limits_{v \in \mathcal{V}_a} s(i, v) > T'_a \\
\text{null} & \text{otherwise}
\end{cases}
\]
Equivalently, we can include the null value as an explicit candidate and retrieve the top-1 value as:
\[
v_a^{\text{pred}} = \arg\max\limits_{v \in \mathcal{V}_a \cup \{v_0^a\}} s(i, v)
\]
In this case, selecting the \(v_0^a\) corresponds to the scenario where none of the specific candidate values surpass the dynamic threshold.

The inference process is illustrated on the right side of Figure~\ref{fig:method}.
In this example, the model retrieves \(v_3^1\), \(v_1^2\), and \(v_2^4\) as the predicted values for attributes \(a_1\), \(a_2\), and \(a_4\), respectively, 
while those for \(a_3\) and \(a_5\) are determined to be empty as their top-scoring values are null.

\section{Experiment Settings}
\label{sec:settings}

\begin{table*}[tb]
\centering
\small
\caption{Statistics of the attribute taxonomies and dataset splits from \datasetxy and \datasetwdc. ``CA Pairs'' refers to category-attribute pairs, ``CAV Tuples'' denotes category-attribute-value tuples, ``PA Pairs'' represents product-attribute pairs, and ``Null Pairs'' indicates product-attribute pairs with null values. ``Excl.'' refers to the test split excluding measurement attributes.}
\label{tab:statistics}

\begin{subtable}[t]{0.7\columnwidth}
    \centering
    \caption{Statistics of the attribute taxonomies.}
    \label{tab:statistics_taxonomy}
    \begin{tabular}{lrr}
    \toprule
    Statistic & Xianyu & WDC \\
    \midrule
    \# Categories    &     8,803 &     5 \\
    \# Attributes    &     3,326 &    24 \\
    \# CA Pairs      &    26,645 &    37 \\
    \# CAV Tuples    & 6,302,220 & 2,297 \\
    \bottomrule
    \end{tabular}
\end{subtable}%
\hfill
\begin{subtable}[t]{1.3\columnwidth}
    \centering
    \caption{Statistics of the datasets.}
    \label{tab:statistics_dataset}
    \begin{tabular}{lrrrrrr}
    \toprule
    \multirow{2}{*}{Statistic} & \multicolumn{3}{c}{\datasetxy} & \multicolumn{3}{c}{\datasetwdc} \\
    \cmidrule(l{0pt}r{5pt}){2-4}\cmidrule(l{5pt}r{0pt}){5-7}
     & Train & Valid & Test & Train & Test & Excl. \\
    \midrule
    \# Products       &   809,528 &  81,699 &  85,024 & 1,066 &   354 &   354 \\
    \# PA Pairs       & 3,584,462 & 358,582 & 458,954 & 8,832 & 2,937 & 2,285 \\
    \# Null Pairs     & 2,345,577 & 228,534 & 272,285 & 3,973 & 1,330 &   916 \\
    \bottomrule
    \end{tabular}
\end{subtable}

\end{table*}

\subsection{Datasets}

To evaluate \taskabbr under the settings described in \S\ref{sec:problem_def}, we compare our proposed method against baselines on both proprietary and public datasets with normalized values\footnote{Other popular benchmarks, such as AE-110k~\cite{xu-etal-2019-scaling} and MAVE~\cite{MAVEQA}, provide only unnormalized values as spans extracted from product profiles, making them unsuitable for our experiments.}.
Table~\ref{tab:statistics} presents statistics of the attribute taxonomies and datasets.

\noindent \textbf{\datasetxy}. This dataset, derived from the e-commerce platform \platform, is constructed to evaluate the scalability and generalization of \taskabbr methods. The platform's attribute taxonomy comprises 8,803 product categories, 26,645 category-attribute pairs, and 6.3 million category-attribute-value tuples. On average, each category is associated with 3 attributes, each attribute has 237 possible values, and there are 716 candidate values per category.

For our experiments, we randomly sampled 1 million product items for training, 10{,}000 for validation, and 10{,}000 for testing. Each item is annotated with a category label and multiple attribute-value pair labels.
Category labels were generated through a multi-step process involving automated classification, seller feedback, and annotator review, during which misclassified samples were discarded.
Attribute-value pair labels were obtained through a multi-stage manual annotation process: a pool of annotators conducted the initial labeling, followed by quality checks and a second round of review, with items reassigned to different annotators after shuffling.

\noindent \textbf{\datasetwdc} \cite{Baumann2024UsingLF}. This dataset contains 1,066 training and 354 test product items across 5 categories, with 8,832 and 2,937 product-attribute pairs (3,973 and 1,330 nulls), respectively.
On average, each category is associated with 7.4 attributes, each attribute has 62 values, and there are 459 attribute-value pairs per category.
We conduct two evaluations: the first on the original test set, and the second on a test split that excludes measurement attributes, which require complex reasoning for unit conversion.

\subsection{Metrics}
\label{sec:metrics}

Since most attributes in the taxonomy are exclusive, i.e., each product can have at most one value per attribute, we evaluate \taskabbr methods using micro-averaged precision@1, recall@1, and F1 score@1.

For each attribute, the ground truth is a set of values \(\mathcal{V}\) from the taxonomy. If the ground truth set is empty (\(\varnothing\)), a correct prediction (True Negative, TN) occurs when the model also predicts an empty set; otherwise, it is a False Positive (FP). When the ground truth set is not empty, the model's top-1 output is a True Positive (TP) if it matches any ground truth value. Predicting an empty set in this case results in a False Negative (FN), while mismatched predictions are both False Positives (FP) and False Negatives (FN), as it simultaneously introduces an error and misses the correct value.
Table \ref{tab:confusion-matrix} summarizes these outcomes\footnote{In prior work \cite{shinzato-etal-2023-unified}, metrics did not account for the FP case, and FP \& FN cases were counted as FP only. We adopt more stringent metrics in this paper.}.
Final precision, recall, and F1 scores are computed by aggregating TP, FP, and FN counts across the dataset, providing a comprehensive performance evaluation.

\begin{table}[tb]
\centering
\small
\caption{Confusion matrix comparing ground truth value set with predicted top-1 value.}
\label{tab:confusion-matrix}
\begin{tabular}{ccl}
\toprule
Label & Prediction & Outcome \\
\midrule
$\varnothing$         & $\varnothing$            & True Negative (TN)  \\
$\varnothing$         & $v$                      & False Positive (FP) \\
$\mathcal{V}$         & $v \in \mathcal{V}$      & True Positive (TP) \\
$\mathcal{V}$         & $\varnothing$            & False Negative (FN) \\
$\mathcal{V}$         & $v' \notin \mathcal{V}$  & FP \& FN \\
\bottomrule
\end{tabular}
\end{table}

\begin{table*}[tb]
\centering
\small
\caption{Performance comparison of classification-, generation-, and retrieval-based methods on \datasetxy{} and \datasetwdc{}. 
"F1 Excl." denotes the F1 score computed excluding measurement attributes (e.g., width and height), which require complex unit normalization reasoning.}
\label{tab:main}
\begin{tabular}{llccccccc}
\toprule
\multirow{2}{*}{Paradigm} & \multirow{2}{*}{Method} & \multicolumn{3}{c}{\datasetxy{}} & \multicolumn{4}{c}{\datasetwdc{}} \\ 
\cmidrule(l{0pt}r{5pt}){3-5}\cmidrule(l{5pt}r{0pt}){6-9}
 &  & Precision & Recall & F1 & Precision & Recall & F1 & F1 Excl. \\ 
\midrule
\multirow{1}{*}{Classification} & BERT-CLS & 50.9 & 50.1 & 50.5 & 68.9 & 12.0 & 20.5 & 23.4 \\ 
\midrule
\multirow{8}{*}{Generation}
    & Llama3.1 (zero-shot) & 29.1 & 46.2 & 35.7 & 56.6 & 60.8 & 58.6 & 64.6 \\ 
    & Llama3.1 (few-shot)  & 31.0 & 51.1 & 38.6 & 76.0 & 74.1 & 75.0 & 79.0 \\ 
    & Llama3.1 (RAG)       & 40.8 & 57.2 & 47.6 & \textbf{78.2} & \textbf{76.3} & \textbf{77.2} & 80.1 \\ 
    & Llama3.1 (fine-tune) & \textbf{86.9} & 82.7 & 84.7 & 57.7 & 60.4 & 59.0 & 64.5 \\ 
    \cmidrule(lr){2-9}
    & Qwen2.5 (zero-shot)  & 42.7 & 55.7 & 48.4 & 51.9 & 60.3 & 55.8 & 60.8 \\ 
    & Qwen2.5 (few-shot)   & 45.8 & 58.6 & 51.4 & 72.2 & 72.3 & 72.2 & 76.2 \\ 
    & Qwen2.5 (RAG)        & 58.3 & 69.1 & 63.2 & 75.1 & 73.4 & 74.2 & 78.3 \\ 
    & Qwen2.5 (fine-tune)  & 84.5 & 79.1 & 81.7 & 54.1 & 60.0 & 56.9 & 61.7 \\ 
\midrule
Retrieval & \method & 85.4 & \textbf{87.1} & \textbf{86.2} & 74.3 & 70.9 & 72.6 & \textbf{80.3} \\ 
\bottomrule
\end{tabular}
\end{table*}

\subsection{Baselines}

We evaluate our retrieval-based method, \method{}, against classification and generation baselines\footnote{Extraction baselines, such as NER and QA models, are not included because (1) they produce unnormalized outputs, requiring an established normalization strategy for fair comparison, and (2) the large-scale \datasetxy{} lacks token-level annotations (e.g., BIO tags) necessary for these methods.}.
For implementation details, refer to Appendix~\ref{sec:implement_details}.

\noindent \textbf{BERT-CLS}. This baseline frames \taskabbr{} as a multi-label classification task, treating each category-attribute-value tuple as an independent label. The model is fine-tuned to predict matches, with label masking applied to exclude irrelevant labels for each category, following~\cite{chen-etal-2022-extreme}. The model outputs a probability distribution over values and selects the highest-probability value for each attribute. If no probability exceeds a specified threshold, the prediction is set to empty.

\noindent \textbf{LLM}. For generation-based baselines, we utilize state-of-the-art open-source LLMs, including Llama3.1-7B~\cite{grattafiori2024llama} and Qwen2.5-7B~\cite{yang2024qwen2technicalreport}. These models are evaluated in zero-shot and few-shot settings using a template adapted from~\cite{Baumann2024UsingLF}, which incorporates the category, attribute, product profile, and detailed value normalization guidelines. We further fine-tune the LLMs using LoRA~\cite{hu2022lora} to improve performance.

\noindent \textbf{RAG}. We implement RAG baselines based on both Qwen2.5 and Llama3.1, using BGE embeddings~\cite{bge_embedding} to retrieve the top-5 most similar product items from the training set. The LLM prompt is augmented with the profiles and structured attribute-value pair outputs of these retrieved examples. For all generation baselines, we apply greedy decoding to ensure reproducibility, and model outputs are formatted in JSON.

\section{Results}

\subsection{Main Results}

Table~\ref{tab:main} presents the performance comparison between our retrieval-based method \method{} and classification- and generation-based baselines. On \datasetxy{}, \method{} achieves the highest F1 score of 86.2\%, surpassing fine-tuned Llama3.1, which obtains 84.7\%. Notably, \method{} excels in recall, achieving 87.1\% compared to Llama3.1's 82.7\%. On \datasetwdc{}, \method{} achieves the highest F1 Excl. score of 80.3\%, which excludes measurement attributes requiring unit normalization reasoning. These results highlight \method{}'s effectiveness and robustness in addressing general \taskabbr{} across diverse datasets.

The classification-based baseline, BERT-CLS, shows the weakest performance on both datasets. It achieves an F1 score of 50.5\% on \datasetxy{}, but its performance drops drastically to 20.5\% on \datasetwdc{}, underscoring the limitations of classification approaches in generalization. One contributing factor is the extreme label sparsity in this formulation: there are over 6.3M category-attribute-value labels, but the training set contains fewer than 3.6M instances, making it difficult for the model to learn a reliable classification head.

Among generation-based methods, performance improves steadily from zero-shot to few-shot, RAG, and fine-tuning on \datasetxy{}. Llama3.1 progresses from 35.7\% (zero-shot) to 38.6\% (few-shot), 47.6\% (RAG), and 84.7\% (fine-tuned), while Qwen2.5 improves from 48.4\% to 51.4\%, 63.2\%, and 81.7\%, respectively. A similar trend holds on \datasetwdc{}, though fine-tuning is less effective due to limited supervision. In this setting, RAG outperforms few-shot prompting for both Qwen2.5 (74.2\% vs. 72.2\%) and Llama3.1 (77.2\% vs. 75.0\%), offering a viable alternative when labeled data is limited. While \method{} substantially outperforms RAG on \datasetxy{} (86.2\% vs. 47.6\%/63.2\%), its advantage is narrower on \datasetwdc{}, where it matches or slightly exceeds RAG in non-measurement attributes (F1 Excl.: 80.3\% vs. 80.1\%/78.3\%).
This likely stems from the limited size of \datasetwdc{}, which constrains the effectiveness of supervised learning approaches, including BERT-CLS, fine-tuned LLMs, and \method{}.

\begin{table}[tb]
        \centering
        \small
        \caption{Inference efficiency comparison on \datasetxy{} (Throughput in samples/second).}
        \label{tab:speed}
        \begin{tabular}{lrr}
        \toprule
        Method & Time (ms) & Throughput \\ 
        \midrule
        BERT-CLS             &  8.6 & 930 \\ 
        \midrule
        Llama3.1 (zero-shot) & 101.3 & 80 \\ 
        Llama3.1 (few-shot)  & 124.8 & 64 \\ 
        Llama3.1 (RAG)       & 137.9 & 58 \\
        Qwen2.5 (zero-shot)  &  84.0 & 95 \\ 
        Qwen2.5 (few-shot)   &  98.4 & 81 \\ 
        Qwen2.5 (RAG)        & 108.3 & 74 \\
        \midrule
        \method              & 12.7 & 630 \\ 
        \bottomrule
        \end{tabular}
\end{table}

\begin{table}[t]
    \centering
    \small
    \caption{F1 scores on product-attribute pairs with normalized vs.\ unnormalized and implicit values.}
    \label{tab:unnorm-implicit-eval}
    \begin{tabular}{lrr}
        \toprule
        Method & Normalized & Unnorm.\ \& Implicit \\ 
        \midrule
        Llama3.1 & 83.2 & 79.4 \\ 
        Qwen2.5  & 82.6 & 78.6 \\ 
        \midrule
        \method  & 87.9 & 82.9 \\ 
        \bottomrule
    \end{tabular}
\end{table}

\subsection{Inference Efficiency}

Table~\ref{tab:speed} presents a comparison of inference efficiency across different \taskabbr{} paradigms under identical conditions, using an unoptimized PyTorch implementation on a single NVIDIA V100 GPU.

\method{} achieves a strong balance between model capacity and efficiency, processing each sample in 12.7~ms and achieving a throughput of 630 samples per second. In contrast, generation-based methods using Llama3.1 and Qwen2.5 exhibit substantially higher inference times (over 100~ms per sample) and lower throughput (below 100 samples per second), primarily due to the overhead of autoregressive decoding and reliance on large-scale models. The BERT-CLS classification baseline offers the highest raw efficiency (8.6~ms, 930 samples per second), but its inability to handle out-of-distribution values and limited generalization capacity reduce its practical applicability.

In summary, among these methods, \method{} provides the best trade-off between identification capability and inference efficiency, making it well-suited for scalable industrial deployment.

\subsection{Analysis}
\label{sec:analysis}

\begin{figure}[t]
    \centering
    \includegraphics[width=0.65\columnwidth]{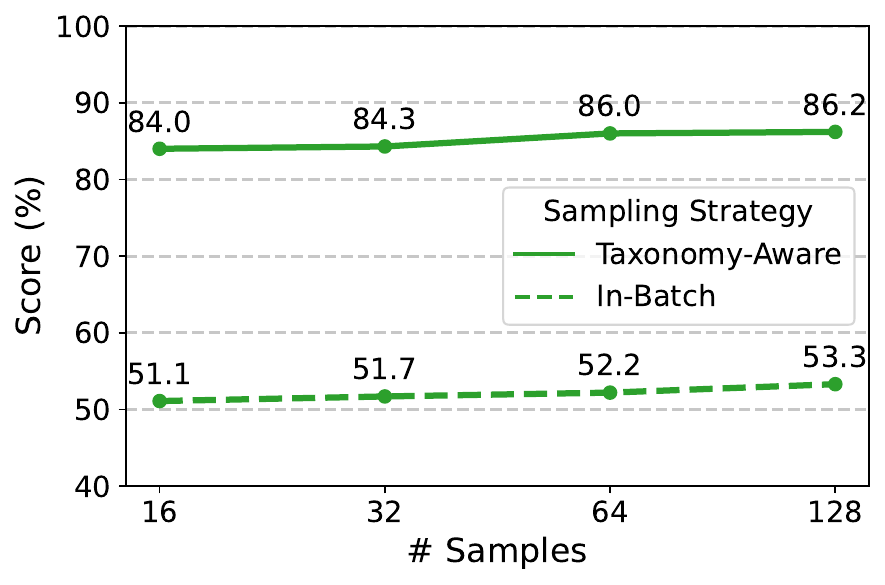}
    \caption{Comparison of negative sampling strategies with increasing number of samples.}
    \label{fig:sampling}
\end{figure}

This section analyzes our method on the \datasetxy{} dataset, selected for its large scale and diverse attribute taxonomy. In contrast, the WDC-PAVE dataset is substantially smaller in both size and coverage, making it insufficient for robust evaluation of scalability and generalization.

\begin{figure*}[t]
    \centering
    \begin{subfigure}[t]{0.33\textwidth}
        \centering
        \includegraphics[width=\textwidth]{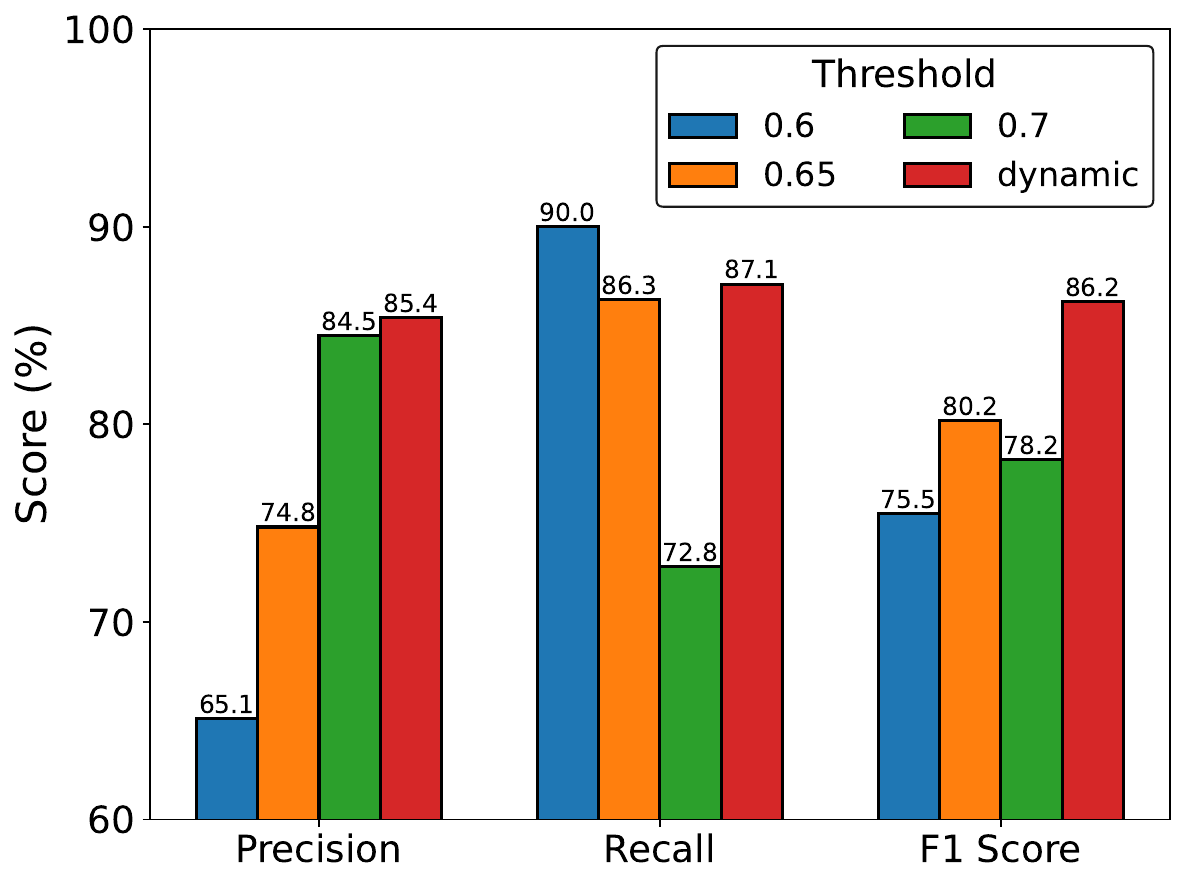}
        \caption{Comparison of inference thresholds.}
        \label{fig:thresholds}
    \end{subfigure}%
    \begin{subfigure}[t]{0.33\textwidth}
        \centering
        \includegraphics[width=\textwidth]{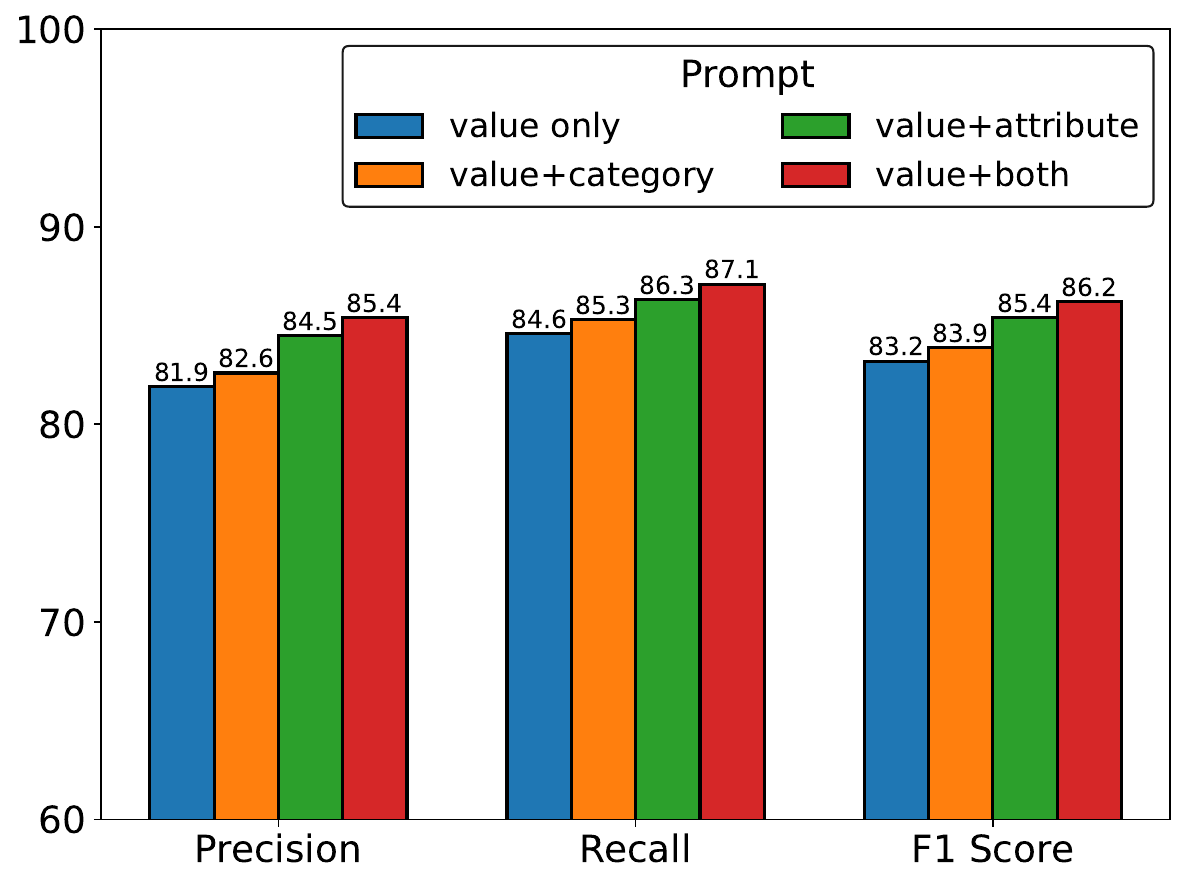}
        \caption{Comparison of prompt templates.}
        \label{fig:prompts}
    \end{subfigure}%
    \begin{subfigure}[t]{0.33\textwidth}
        \centering
        \includegraphics[width=\textwidth]{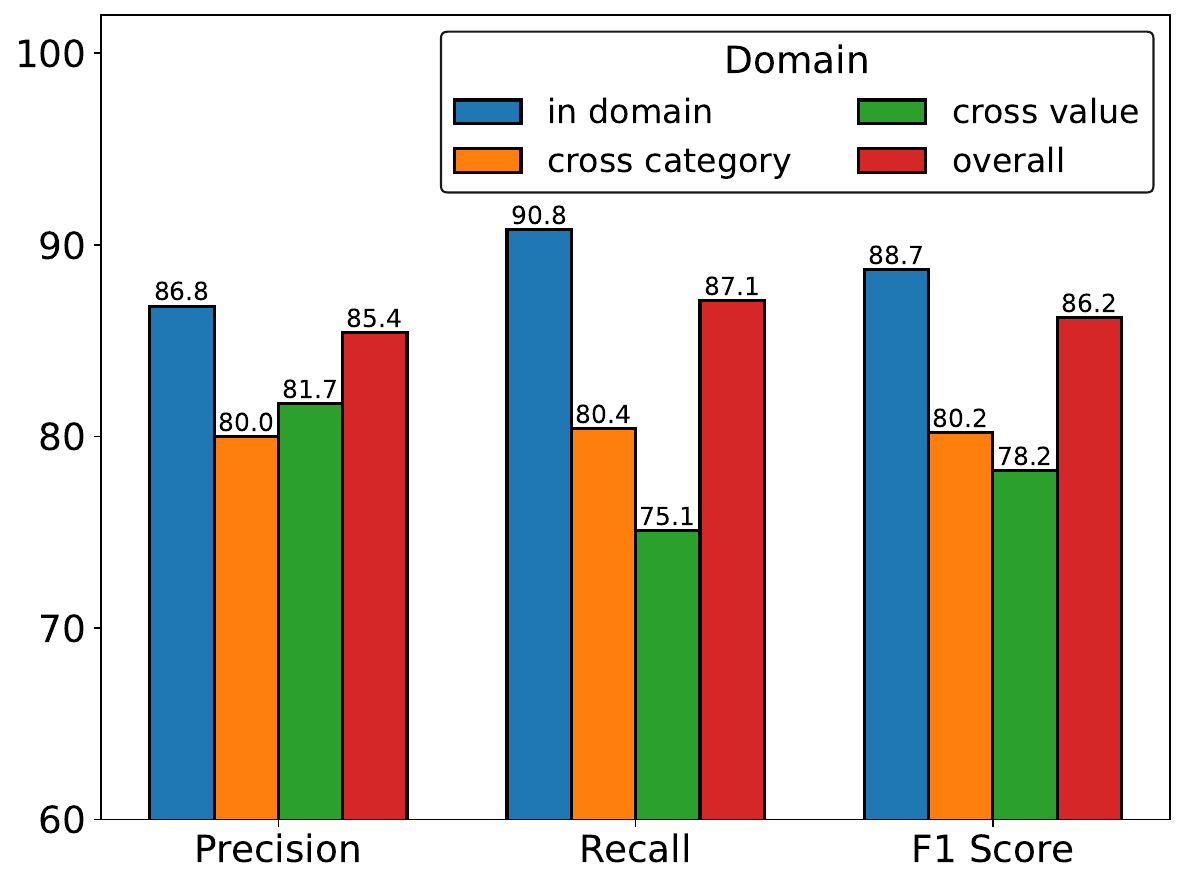}
        \caption{Comparison of data domains.}
        \label{fig:transfer}
    \end{subfigure}
    \caption{Performance analysis across inference thresholds, prompt templates, and data domains.}
    \label{fig:combined_analysis}
\end{figure*}

\noindent \textbf{Robustness Evaluation on Diverse Values.}
To better assess model robustness under various conditions, we partition the test set into three subsets based on the nature of the ground truth: (1) \textit{normalized values}, appear verbatim in the product profile; (2) \textit{unnormalized or implicit values}, which either appear in a lexically varied form, or must be inferred; and (3) \textit{null values}, where the attribute value is marked as unavailable. In our \datasetxy{} dataset, normalized values account for approximately 15.2\%, unnormalized or implicit values for 27.2\%, and null values for 57.6\%.

Table~\ref{tab:unnorm-implicit-eval} reports the F1 scores on the normalized and unnormalized/implicit subsets. \method{} achieves the highest performance on both subsets, with an F1 score of 87.9\% on normalized values and 82.9\% on unnormalized or implicit values. Compared to the fine-tuned LLM baselines, \method{} demonstrates stronger ability to recognize standardized values and better robustness to lexical variation and implicit inference.

\noindent \textbf{Impact of Taxonomy-Aware Negative Sampling.}
Figure~\ref{fig:sampling} compares the proposed taxonomy-aware negative sampling (\S\ref{sec:contrastive_training}) with in-batch negative sampling across varying sample sizes. As the number of sampled values increases, the F1 score improves consistently, aligning with findings from \cite{pmlr-v119-chen20j}. Using in-batch sampling as the baseline, the model achieves an F1 score of 53.3\% with a sample size of 128. In contrast, taxonomy-aware sampling yields substantial improvements, boosting the F1 score from 84.0\% to 86.2\% as the sample size grows from 16 to 128.

These results demonstrate that taxonomy-aware sampling provides more effective supervision, encouraging the model to distinguish between fine-grained, semantically similar values within the same category and attribute. In contrast, in-batch sampling often yields random negatives that are less relevant and frequently trivial, limiting the efficacy of contrastive learning for \taskabbr.
For example, in Figure~\ref{fig:tasks}, negatives such as \texttt{iPhone 12 Pro} or \texttt{iPhone 13 Pro Max}, which belong to the \texttt{model} attribute under the \texttt{phone} category, provide more challenging and informative supervision than random values such as \texttt{L} for T-shirt size.

\noindent \textbf{Comparison of Dynamic and Static Thresholds.}  
Figure~\ref{fig:thresholds} compares our dynamic thresholding approach (\S\ref{sec:adaptive_inference}) with static thresholds of 0.6, 0.65, and 0.7, selected via validation.
The dynamic threshold achieves the highest F1 (86.2\%), outperforming the static baselines (75.5\%, 80.2\%, and 78.2\%).
Static thresholds exhibit the typical precision–recall trade-off: higher thresholds increase precision (65.1\%→84.5\%) but reduce recall (90.0\%→72.8\%). In contrast, the dynamic threshold balances precision (85.4\%) and recall (87.1\%) without manual tuning.
This gain stems from a scalable design: instead of relying on fixed, hand-tuned cutoffs per category-attribute pair, our method learns null value embeddings whose similarity scores act as adaptive thresholds.

\noindent \textbf{Performance Gains from Context-Rich Prompts.}
The influence of varying value prompt templates on the \taskabbr task is shown in Figure~\ref{fig:prompts}. Using only the value as a prompt achieves an F1 score of 83.2\%. Adding category information raises the F1 score to 83.9\%, while incorporating attribute information further improves it to 85.4\%. The most comprehensive template, combining category, attribute, and value information (i.e., \texttt{A \{category\} with \{attribute\} being \{value\}}), achieves the highest F1 score of 86.2\%. These results are consistent with prior work~\cite{pmlr-v139-radford21a}, highlighting that context-rich prompts enhance the model's discriminative performance.

\noindent \textbf{Zero-Shot Generalization Across Data Domains.}  
Figure~\ref{fig:transfer} presents zero-shot transfer results on unseen categories and values. The in-domain split achieves an F1 score of 88.7\%, while cross-category and cross-value splits decline to 80.2\% and 78.2\%, respectively, reflecting the challenges of adapting to evolving attribute taxonomies in out-of-distribution domains. Despite this, \method maintains a strong overall F1 of 86.2\%, demonstrating robust generalization. The model's generalization relies on the shared textual encoder’s semantic understanding and the retrieval-based approach’s capacity to leverage these embeddings for zero-shot matching. Such adaptability is critical for dynamic e-commerce platforms, where new products and attribute-value pairs continuously emerge, reducing the need for frequent retraining and lowering maintenance costs.

\section{Conclusion}
In this work, we present \method, a novel retrieval-based approach for \taskabbr. By formulating \taskabbr\ as an information retrieval problem, \method enables the inference of implicit values, generalization to OOD values, and the production of normalized outputs. Building on this framework, \method employs contrastive training with taxonomy-aware sampling and adaptive inference with dynamic thresholds to enhance retrieval performance and scalability.

Comprehensive experiments on proprietary and public datasets demonstrated \method's superiority over classification- and generation-based baselines. Notably, \method achieved an F1 score of 86.2\% on the large-scale \datasetxy dataset. Our efficiency analysis further highlighted its advantage, achieving significantly faster inference speeds than generation-based methods. Beyond these experimental results, \method has been successfully deployed on the real-world e-commerce platform \platform, processing millions of product listings daily and seamlessly adapting to dynamic attribute taxonomies, making it a practical solution for large-scale industrial applications.

\section{Limitations}

\method assumes access to a predefined attribute taxonomy, which serves as a foundation for accurate value identification. While this setup is realistic in many e-commerce platforms, where dedicated systems and teams maintain evolving taxonomies, it does require ongoing manual updates to incorporate new categories, attributes, and values. Automating product attribute mining remains an open area for future research~\cite{10.1145/1147234.1147241, 10.1145/3485447.3512035, xu-etal-2023-towards}.

Another limitation is that \method currently operates on textual product profiles and does not incorporate multimodal information, such as images or videos. Multimodal inputs could provide complementary signals for attributes that are difficult to infer from text alone (e.g., color, material, or shape). Extending the method to support multimodal input could further improve its coverage and accuracy in practical applications.

A further limitation of \method{}, as a retrieval-based approach, is its difficulty in handling measurement attributes that require unit conversion or numerical reasoning. To address this, future work could explore hybrid methods that combine \method{}'s retrieval strengths with the reasoning ability of LLMs to improve performance in such cases.


\bibliography{anthology,custom}

\appendix

\section{Implementation Details}
\label{sec:implement_details}

For both the BERT-CLS baseline and our \method method, we utilize pre-trained RoBERTa-base~\cite{Liu2019RoBERTaAR, cui-etal-2020-revisiting} as the backbone. 
For \method, we augment the model with a linear projection head to map the embedding dimension to 256. For each product-attribute pair, we sample up to 128 values for the contrastive learning setup, including a null value, an optional positive value (if present for the attribute), and negative values sampled from the value set with the same category and attribute. In the case of the null value, we replace the value slot in the prompt template with \texttt{null} (e.g., \texttt{A phone with capacity being null.}).

The temperature parameter is fixed at 0.05. This choice is motivated by both prior research (e.g., SimCLR~\cite{pmlr-v119-chen20j} and MoCo~\cite{9157636}) and empirical tuning on our validation set. The encoder is fine-tuned using the AdamW optimizer, with a batch size of 32, a learning rate of $2\times10^{-5}$, and a maximum of 5 epochs.

For LLM fine-tuning, we employ LoRA~\cite{hu2022lora} for efficient adaptation. The core hyperparameters are as follows: 3 training epochs, batch size of 128, AdamW optimizer, maximum learning rate of $5\times10^{-5}$, 1\% warmup steps, cosine learning rate scheduler, LoRA rank of 8, LoRA alpha of 16, and LoRA dropout rate of 0.1.

All hyperparameters and model checkpoints are selected to maximize the F1 score on the validation set. For further details, please refer to our codebase: \url{https://github.com/SuYindu/TACLR}.

\section{Deployment}

The proposed \method has been successfully integrated into key functionalities of the e-commerce platform \platform, including product listing, search, recommendation, and price estimation. The system is designed to be highly scalable, efficiently processing millions of products daily.

During the product listing process, \method automatically identifies attribute–value pairs from user-provided titles and descriptions. This automation significantly reduces manual effort and errors, while enhancing the quality of structured product information.

For product search, the improved structured information directly supports more effective lexical retrieval and enables structured search, where items are filtered based on attribute values. This leads to more accurate matching with user queries. In addition, the enriched product features enhance the quality of personalized recommendations.

In the context of price estimation, \method identifies key attribute values that influence pricing, enabling more accurate price predictions. This functionality provides sellers and buyers with reliable, market-aligned information in the context of second-hand transactions.


\end{document}